\documentclass[10pt,twocolumn]{article}
\usepackage[T1]{fontenc}
\usepackage{times}
\usepackage[margin=0.75in,columnsep=0.25in]{geometry}
\usepackage{amsmath,amssymb,booktabs,tabularx,microtype,graphicx,xcolor}
\usepackage{algorithm}
\usepackage[noend]{algpseudocode}
\usepackage[numbers,sort&compress]{natbib}
\usepackage[hidelinks]{hyperref}
\usepackage[font=footnotesize,labelfont=rm,skip=3pt]{caption}

\newcommand{\vnWriters}{42}
\newcommand{\vnLeft}{21}
\newcommand{\vnRight}{21}
\newcommand{\vnImages}{336}
\newcommand{\vnShared}{252}
\newcommand{\vnSpecific}{84}
\newcommand{\vnTrunc}{13}

\newcommand{\vHcA}{77.4}

\newcommand{\vHcAsd}{4.3}
\newcommand{\vHcAuc}{0.85}

\newcommand{\vStyleA}{55.4}

\newcommand{\vStrokeA}{60.8}

\newcommand{\vCombA}{75.5}

\newcommand{\vScnnA}{65.8}

\newcommand{\vStrokeAmin}{47.6}
\newcommand{\vStrokeAmax}{73.8}
\newcommand{\vDeepArange}{69.5--72.0}

\newcommand{\vHcP}{0.001}
\newcommand{\vStyleP}{0.266}
\newcommand{\vStrokeP}{0.090}
\newcommand{\vCombP}{0.002}
\newcommand{\vScnnP}{0.020}

\newcommand{\vVitP}{0.004}
\newcommand{\vPhotoP}{0.025}
\newcommand{\vPaperP}{0.095}
\newcommand{\vPhotoA}{66.4}
\newcommand{\vPaperA}{61.5}

\newcommand{\vBlkHinge}{62.6}
\newcommand{\vBlkDir}{51.3}
\newcommand{\vBlkOrient}{58.2}
\newcommand{\vBlkEnd}{55.7}
\newcommand{\vBlkRates}{47.0}
\newcommand{\vBlkTurn}{60.8}

\newcommand{\vLearnHHc}{70.7}

\newcommand{\vEvHcOne}{72.1}

\newcommand{\vHcCminusD}{+1.0}
\newcommand{\vHcBminusE}{$-$7.6}
\newcommand{\vStrokeCminusD}{$-$6.2}
\newcommand{\vStrokeBminusE}{$-$5.8}

\newcommand{\vHcAdj}{64.5}

\newcommand{\vStrokeAdj}{60.0}
\newcommand{\vCombAdj}{64.0}
\newcommand{\vScnnAdj}{51.1}

\newcommand{\vDeepAdjRange}{52.0--61.0}
\newcommand{\vStrokeAdjAllP}{0.084}

\newcommand{\vStrokeWid}{100.0}
\newcommand{\vHcWid}{73.0}
\newcommand{\vDeepWidRange}{93.7--94.4}
\newcommand{\vStrokeLeaky}{94.2}
\newcommand{\vStrokeParaSplit}{59.1}
\newcommand{\vDeepLeakyRange}{91.1--95.4}
\newcommand{\vDeepParaRange}{67.2--70.0}

\newcommand{\vDupShiftMin}{$-$2.2}
\newcommand{\vDupShiftMax}{+3.5}
\newcommand{\vStrokeMirOrig}{60.8}
\newcommand{\vStrokeMir}{48.2}
\newcommand{\vStrokeMirFlip}{39}

\newcommand{\vGreyFtA}{71.5}
\newcommand{\vGreyFtLo}{59.6}
\newcommand{\vGreyFtHi}{82.6}

\newcommand{\vGreyFtEA}{69.0}

\newcommand{\vGreyFtAdj}{54.9}
\newcommand{\vGreyFtAdjLo}{43.7}
\newcommand{\vGreyFtAdjHi}{65.7}
\newcommand{\vGreyFtC}{72.0}
\newcommand{\vGreyFtB}{70.6}

\newcommand{\vPaperLevP}{0.289}

\newcommand{\vBlockFracL}{79}
\newcommand{\vBlockFracR}{78}
\newcommand{\vBlockFracP}{0.464}
\newcommand{\vTiltL}{2.06}
\newcommand{\vTiltR}{1.82}
\newcommand{\vTiltP}{0.55}
\newcommand{\vSigHinge}{0}
\newcommand{\vNHinge}{144}
\newcommand{\vSigSlant}{3}
\newcommand{\vNSlant}{9}
\newcommand{\vSigOrient}{0}
\newcommand{\vNOrient}{12}

\newcommand{\vSlantBackG}{$-$0.73 to $-$0.78}
\newcommand{\vSlantBackQ}{0.059}

\newcommand{\vSlantBackL}{22.2}
\newcommand{\vSlantBackR}{24.1}

\newcommand{\vInkConG}{$-$0.70}
\newcommand{\vInkConP}{0.028}
\newcommand{\vInkDensP}{0.110}

\newcommand{\vOutlShiftMin}{$-$3.8}
\newcommand{\vOutlShiftMax}{+2.7}
\newcommand{\vTruncShiftMin}{$-$4.3}
\newcommand{\vTruncShiftMax}{+0.1}
\newcommand{\vPhotoAuc}{0.75}

\newcommand{\vSpStrokeHc}{0.24}
\newcommand{\vAuditMaxDiff}{102}

\newcommand{\affl}[1]{{\small\itshape #1}}
\begin{document}
\title{Writer-independent handedness identification from offline Bangla handwriting under controlled acquisition}
\author{Tanzim Hasan Prappo\\[2pt]
\affl{Dept.\ of Computer Science and Engineering}\\\affl{Shahjalal University of Science and Technology}\\\affl{Sylhet, Bangladesh}\\[1pt]
\small\href{mailto:2021331006@student.sust.edu}{2021331006@student.sust.edu}
\and
Sayeed Hassan\\[2pt]
\affl{Dept.\ of Computer Science and Engineering}\\\affl{Shahjalal University of Science and Technology}\\\affl{Sylhet, Bangladesh}\\[1pt]
\small\href{mailto:2021331044@student.sust.edu}{2021331044@student.sust.edu}
\and
Sayem Ali\\[2pt]
\affl{Dept.\ of Computer Science and Engineering}\\\affl{Shahjalal University of Science and Technology}\\\affl{Sylhet, Bangladesh}\\[1pt]
\small\href{mailto:2023331036@student.sust.edu}{2023331036@student.sust.edu}
\and
Sukhendra Nath Dhar\\[2pt]
\affl{Dept.\ of Computer Science and Engineering}\\\affl{Shahjalal University of Science and Technology}\\\affl{Sylhet, Bangladesh}\\[1pt]
\small\href{mailto:2021331024@student.sust.edu}{2021331024@student.sust.edu}
\and
M. Shahidur Rahman\\[2pt]
\affl{Dept.\ of Computer Science and Engineering}\\\affl{Shahjalal University of Science and Technology}\\\affl{Sylhet, Bangladesh}\\[1pt]
\small\href{mailto:rahmanms@sust.edu}{rahmanms@sust.edu}}
\date{}
\twocolumn[
\maketitle
\vspace{0.5em}
\begin{abstract}
Handedness is a soft-biometric cue for forensic document examination and writer profiling, yet reported accuracies from offline handwriting rarely keep writers, texts and image acquisition apart, so models may recognise the writer or capture conditions rather than the writing hand. Such controlled evidence is missing for Bangla. Using our BanglaHand dataset of \vnWriters{} writers (\vnLeft{} left- and \vnRight{} right-handed), who each wrote six shared prompts and two own texts under controlled collection and photography, we evaluate handedness identification with all images of each test writer held out, whole-writer permutation tests and photometric controls. An ink-gradient descriptor identified the handedness of unseen writers with \vHcA\% balanced accuracy (95\% writer-bootstrap interval 67.7--86.5\%; $p=\vHcP$) and frozen ImageNet embeddings with \vDeepArange\%. Part of this signal is stroke darkness: in a post hoc analysis, right-handed writers' strokes were darker at their centres for equal stroke width, consistent with higher pen pressure and not explained by lighting drift. With ink density and brightness regressed out, the descriptor still reached \vHcAdj\% ($p=0.027$), whereas writing-only shape features were weaker (\vStrokeA\%; $p=\vStrokeP$). Handedness is thus identifiable from offline Bangla handwriting under controlled acquisition, and photometric controls are essential to interpret such cues.
\end{abstract}
\vspace{1.5em}
]
\section{Introduction}
Automatic handwriting analysis has sought to infer soft-biometric attributes such as gender, age and handedness \cite{liwicki2011gender,bouadjenek2016soft}. Left-handed writers may hold the pen, rotate the page and move across a line differently from right-handed writers, so offline images could retain traces in slant, stroke direction or stroke endings. Reported handedness accuracies range from F-measures of about 70\% \cite{almaadeed2016handedness} to above 99\% \cite{rahmanian2021handedness} (see also \cite{morera2018handedness}), but such numbers are hard to interpret unless two conditions are controlled.

First, handwriting identifies its writer very well \cite{bulacu2007text}. If paragraphs of one person appear in training and test data, a classifier can recognise the person and inherit the label. Second, any systematic difference in how the two groups were collected or photographed is a shortcut that learned models exploit readily \cite{geirhos2020shortcut}. Neither problem is visible in an accuracy figure alone.

Bangla script, with its continuous headstroke (\emph{matra}), hanging vowel signs and dense conjuncts, has public offline handwriting datasets \cite{mridha2021banglawriting}, but none for handedness. We use the \emph{BanglaHand} dataset \cite{banglahand2026data}, in which every writer wrote six shared prompts and two own texts, and ask whether handedness can be predicted for writers absent from training, whether the prediction survives controls for acquisition, and which stroke properties carry it.

The contributions are: (i) a writer-independent protocol with matched text conditions, repeated cross-validation, whole-writer permutation tests, acquisition controls and sensitivity analyses, specified in advance; (ii) writing-only stroke features that exclude grey levels, ink amount and framing, with mirror diagnostics of their direction dependence; and (iii) evidence that handedness is identifiable for unseen writers under controlled collection and acquisition, partly through stroke darkness that is not explained by photography, with a decomposition into darkness, gradient texture and writing-only shape.

\section{Data}
\label{sec:data}
\begin{figure*}[t]
\centering
\includegraphics[width=0.92\textwidth]{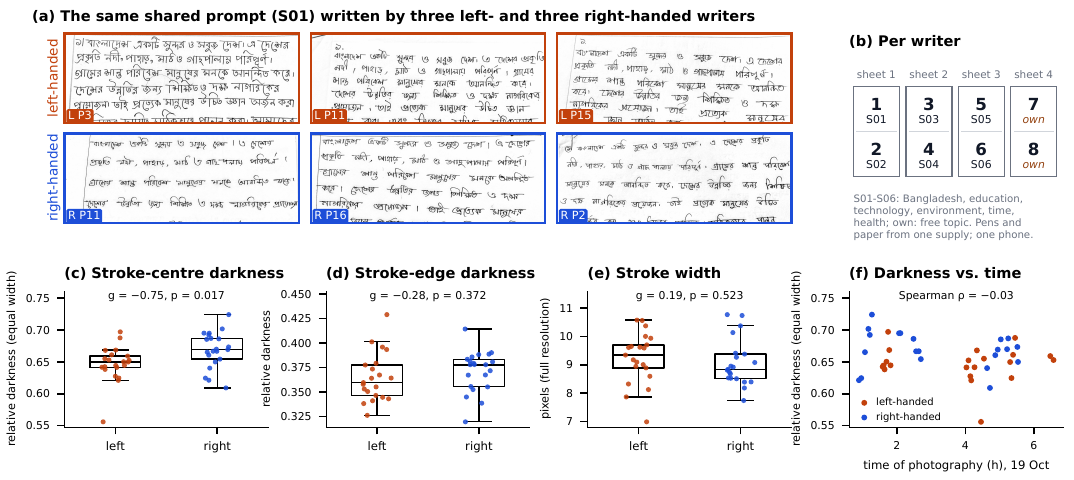}
\caption{The BanglaHand dataset \cite{banglahand2026data} and the stroke-darkness cue. (a) The shared prompt S01 written by three left-handed (orange) and three right-handed (blue) writers; first lines of the text block; for display only, uneven lighting was flattened and contrast stretched (all models use the unmodified photographs). (b) Each writer filled four sheets with two paragraphs each: six shared prompts and two texts of their own choice. (c)--(f) Pseudo-pressure measures (writer means over the six shared paragraphs; post hoc; Section~\ref{sec:photo}): relative darkness of stroke centres at equal stroke width (c) and of stroke edges (d), stroke width (e), and centre darkness against the time of photography (f).}
\label{fig:dataset}
\end{figure*}

BanglaHand \cite{banglahand2026data} was collected as follows. Writers handwrote eight paragraphs on four sheets with pens and paper from one common supply, handed out to left- and right-handed writers alike in mixed writing sessions: six shared Bangla prompts of 49--54 words (S01--S06: Bangladesh, education, technology, environment, time, health) and two paragraphs on topics of their own choice (Fig.~\ref{fig:dataset}). Each writer's label, the group under which their pages were collected, was established by self-report, confirmed with a short handwriting screening task and observed during collection; all participants wrote only with the hand of their group. All left-handed writers were university students recruited at Shahjalal University of Science and Technology (SUST); right-handed writers were primarily SUST university students along with a few local college students (1--2 years apart in age, at a comparable level of education). All writers were 18--25 years old and about 70\% were male, with a similar gender ratio in both groups (gender is not part of the per-writer metadata). Each paragraph was photographed separately with one smartphone (4080$\times$2296 pixels) in a single session, the two groups alternating in blocks of two to nine writers.

Paragraph identity was verified by reading every image, and a geometric screen (SIFT \cite{lowe2004sift} with RANSAC) found no shared pages among writers (at most \vAuditMaxDiff{} inliers between different writers); a raw folder that re-photographed a released writer's pages and a raw right-handed folder excluded by the data preparer as a writer-level outlier in stroke width and ink measures (re-included in a sensitivity analysis, Section~\ref{sec:photo}) are not part of the dataset. The study uses \vnWriters{} writers (\vnLeft{} left, \vnRight{} right) and \vnImages{} images (\vnShared{} shared, \vnSpecific{} writer-specific); \vnTrunc{} images cut the last line at the frame edge. The dataset documentation \cite{banglahand2026data} gives full collection and curation details.

\section{Methods}
\begin{figure*}[t]
\centering
\includegraphics[width=0.9\textwidth]{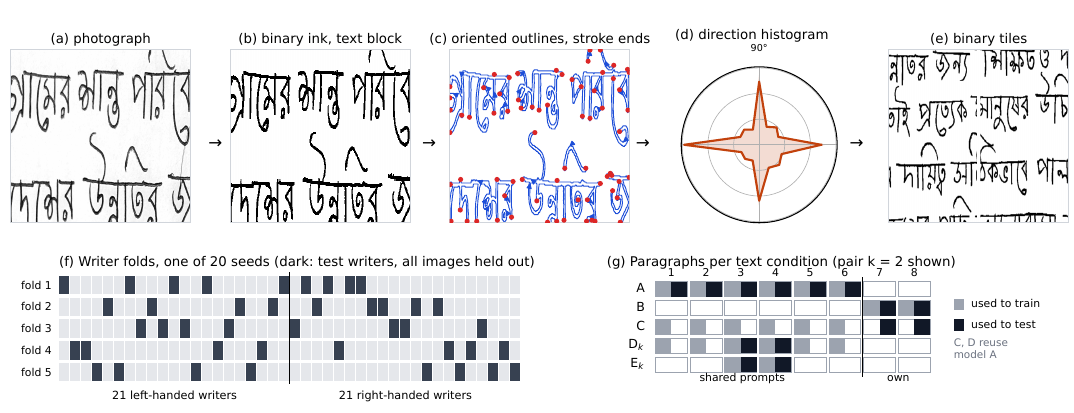}
\caption{Methodology. (a)--(e) From the photograph to the binary ink of the text block, its consistently oriented outlines and stroke ends (red), the 360$^\circ$ contour-direction histogram of this paragraph (one of the STROKE blocks) and binary, stroke-width-normalised tiles for the Stroke-CNN; (a)--(c) show the same image window; (a) is display-corrected as in Fig.~\ref{fig:dataset}. HC-183, the frozen ImageNet models and the photo controls use the grey photograph (a); STROKE and the Stroke-CNN use only (b)--(e). (f) Writer-independent folds of one of the 20 repetitions. (g) Paragraphs used for training and testing in each text condition.}
\label{fig:method}
\end{figure*}

\subsection{Representations}
\textbf{HC-183} (primary; pipeline in Fig.~\ref{fig:method}). The image is resized to 768 pixels width; pixels more than 17 grey levels darker than a Gaussian illumination estimate ($\sigma=17$) form an ink mask after a $2\times2$ opening. On a $4\times4$ grid each cell contributes a 9-bin magnitude-weighted histogram of \emph{unsigned} gradient orientation and its ink density (160 values); row and column ink projections give 18 statistics and connected-component areas 5 quantiles. Related to HOG, it cannot distinguish a stroke from its mirror image and carries ink amount.

\textbf{Style-6} (baseline): slant, stroke width, ink density, horizontal and vertical stroke energy, and row-profile variation.

\textbf{STROKE} (writing-only, 208 dimensions). The ink mask is computed at 1,536 pixels width and cropped to the text block, so margins and framing do not enter; no grey level is used afterwards. Outlines are traced with a fixed orientation (ink on the left) and resampled every 2 pixels; legs of length $L=2w$, where $w$ is the writer's mean stroke width, define at each point the outgoing direction $\phi_1\in[0,2\pi)$ and the incoming direction $\phi_2$. STROKE concatenates the 16-bin histogram of $\phi_1$ (contour direction over the full circle), the $12\times12$ joint histogram of $(\phi_1,\phi_2)$ (contour hinge \cite{bulacu2007text}), a 12-bin histogram of turning angles, a 12-bin histogram of the direction in which skeleton strokes end, 12-bin stroke-orientation and 9-bin slant histograms of skeleton pixels, and endpoint, junction and hole rates per unit stroke length. Every histogram sums to one. Because directions are kept over $360^\circ$ along oriented outlines, mirroring an image permutes the direction and hinge bins, whereas turning is mirror-invariant; a unit test confirms both properties and exact invariance to a brightness shift. Algorithm~\ref{alg:stroke} summarises the extraction.

\begin{algorithm}[t]
\caption{STROKE features of one paragraph image}
\label{alg:stroke}
\footnotesize
\begin{algorithmic}[1]
\Require grey image $I$
\State $I \gets$ resize $I$ to width 1,536; $B \gets$ Gaussian$(I,\sigma{=}34)$
\State $M \gets (B-I>17)\wedge(B>120)\wedge(I<205)$; open with $2{\times}2$; drop components $<12$ px
\State $M \gets$ crop $M$ to rows/columns holding the text block \Comment{no margins}
\State $S \gets$ skeleton$(M)$; $w \gets 2\,\overline{\mathrm{EDT}(M)}|_S$; $L \gets \max(4, 2w)$
\For{each outline $c$ of $M$, oriented with ink on the left}
  \State resample $c$ every 2 px; at each point $x$ take legs to $x_{+L}$ and $x_{-L}$
  \State $\phi_1 \gets \angle(x_{+L}-x)$, $\phi_2 \gets \angle(x_{-L}-x)$ in $[0,2\pi)$; turning $\tau \gets \phi_1-\phi_2-\pi$
\EndFor
\State $h_{\mathrm{dir}} \gets$ hist$_{16}(\phi_1)$; $h_{\mathrm{hinge}} \gets$ hist$_{12\times12}(\phi_1,\phi_2)$; $h_{\mathrm{turn}} \gets$ hist$_{12}(\tau)$
\State $h_{\mathrm{end}} \gets$ hist$_{12}$(direction from $S$ interior to each end point, path length $L$)
\State $h_{\mathrm{ori}}, h_{\mathrm{slant}} \gets$ hist$_{12}$, hist$_9$ of stroke orientation on $S$ (slant: within $45^\circ$ of vertical)
\State $r \gets$ end points, junctions and holes per 1,000 skeleton pixels
\State \Return $[h_{\mathrm{dir}}, h_{\mathrm{hinge}}, h_{\mathrm{turn}}, h_{\mathrm{end}}, h_{\mathrm{ori}}, h_{\mathrm{slant}}, r]$, each histogram normalised to sum 1
\end{algorithmic}
\end{algorithm}

\textbf{CNN/ViT.} Frozen ImageNet ResNet-18 \cite{he2016resnet}, EfficientNet-B0 \cite{tan2019efficientnet} and ViT-B/16 \cite{dosovitskiy2021vit} embed 224-pixel tiles of the grey photograph at 1,568 pixels width (tiles with at least 1\% ink, averaged). \textbf{Stroke-CNN} uses the frozen ResNet-18 on black-on-white tiles of the binary ink rescaled to a stroke width of 3 pixels. \textbf{Fine-tuned ResNet-18} (post hoc baseline) trains layer4 and a new two-way head on the same grey tiles of the training writers' shared paragraphs in each outer fold (4 epochs, AdamW, class-balanced loss, random shifts and scaling); writers are classified end to end by the mean tile probability or by the linear SVM on its averaged features. \textbf{Combined} concatenates HC-183, Style-6 and STROKE.

\textbf{Acquisition controls.} Seven grey-level statistics of the whole photograph (\emph{photo intensity}) and of paper pixels outside the dilated ink (\emph{paper only}).

\subsection{Classifiers, design and statistics}
Features are standardised inside the training folds. The primary classifier is a linear SVM with fixed $C=0.1$; logistic regression ($C$ from four values) and random forests (depth from three values) are tuned by four-fold cross-validation over training writers only \cite{cawley2010selection}. A writer is called left-handed if the mean paragraph SVM margin is $\geq0$ (or the mean probability $\geq0.5$). Outer evaluation is stratified five-fold cross-validation over writers repeated with 20 seeds; every image of a test writer is held out. Text conditions (Fig.~\ref{fig:method}): A, train and test on paragraphs 1--6; B, on 7--8; C, model A applied to 7--8; D$_{1..3}$, model A applied to one shared pair; E$_{1..3}$, train and test on one shared pair. The endpoint is writer balanced accuracy (BA) averaged over repetitions.

Inference uses 999 whole-writer label permutations with class counts preserved, regenerating the folds and refitting all 100 models each time; $p=(b+1)/1000$ \cite{phipson2010permutation}. Sensitivity analyses exclude the \vnTrunc{} truncated images or re-include the outlier folder excluded by the data preparer. An adjustment regresses ink density and the seven photo-intensity statistics out of every feature inside each training fold; post hoc, it was repeated with the time of photography (linear and quadratic terms) as covariates, alone and together with ink density and photo intensity. Pseudo-pressure measures \cite{ammar1986pressure,franke2004ink}, relative darkness $(B-I)/B$, with $I$ the grey level and $B$ the local paper level, on stroke centrelines and edges, centreline darkness at a fixed stroke width (8.4 px) and stroke width, were computed at full resolution, as proxies on compressed smartphone images, and compared between groups with 10,000 permutations (post hoc). The mirror test applies models fitted on original images to left--right mirrored test images; a mirror-augmented STROKE SVM, trained on original and mirrored ink of the training writers, removes access to stroke direction, and the mirror effect (BA on original minus BA on mirrored test images) is tested with 999 whole-writer permutations (post hoc). Intervals are 95\% percentile intervals from 2,000 class-stratified bootstrap resamples of writers applied to the stored predictions of all repetitions (conditional on the fitted models). HC-183 and STROKE in condition A are the primary analyses, specified before any model was fitted on these photographs; all other $p$-values are secondary and not corrected for multiplicity. Writer-level differences of descriptors and of every STROKE bin use 10,000 label permutations with Benjamini--Hochberg control \cite{benjamini1995fdr} within each block (post hoc, descriptive; not used for model selection). Analyses marked post hoc were added after the primary results.

\section{Results}
\subsection{Writer-independent prediction}
\begin{figure*}[t]
\centering
\includegraphics[width=0.8\textwidth]{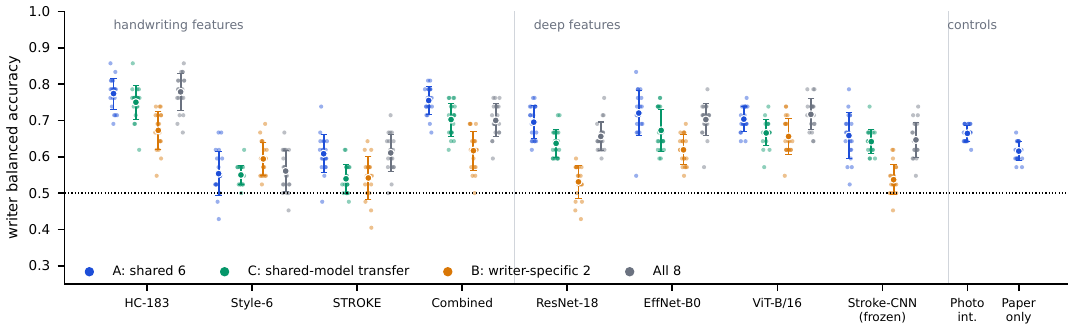}
\caption{Writer balanced accuracy (linear SVM). Small points are repetitions, large points and bars their mean and s.d.; dotted line: chance.}
\label{fig:main}
\end{figure*}

Table~\ref{tab:main} and Fig.~\ref{fig:main} report all representations. With the linear SVM in condition A, HC-183 reached \vHcA\% (s.d.\ \vHcAsd; AUC \vHcAuc; $p=\vHcP$), Combined \vCombA\% ($p=\vCombP$) and the frozen ImageNet embeddings \vDeepArange\% (all $p\leq\vVitP$). The frozen Stroke-CNN reached \vScnnA\% ($p=\vScnnP$). Fine-tuning ResNet-18 on the grey tiles (post hoc, not permutation-tested) did not improve on the frozen embeddings in condition A (\vGreyFtA\% with the SVM on its features, interval \vGreyFtLo--\vGreyFtHi\%; \vGreyFtEA\% end to end) but transferred better to own texts (C \vGreyFtC\%, B \vGreyFtB\%). The writing-only STROKE features were weaker, \vStrokeA\% (range \vStrokeAmin--\vStrokeAmax\% over repetitions; $p=\vStrokeP$), and Style-6 was at chance (\vStyleA\%, $p=\vStyleP$).

\begin{table*}[t]
\caption{Writer balanced accuracy (\%, mean$\pm$s.d.\ over 20 repetitions of five-fold writer cross-validation; \vnWriters{} writers). A: shared prompts (linear SVM, logistic regression, random forest); C: model A applied to the writers' own texts; B: trained and tested on own texts; All~8: all paragraphs (C, B and All~8 with the SVM). 95\% CI: writer-bootstrap interval of the A: SVM value (2,000 resamples, conditional on the fitted models). AUC: within-fold, condition A, SVM. ResNet-18 (fine-tuned): layer4 and head trained on the grey tiles of condition A within each training fold (post hoc); all columns use its features. Controls use grey levels only.}
\label{tab:main}
\centering\footnotesize
\begin{tabular}{lcccccccc}
\toprule
Representation & A: SVM & 95\% CI & A: LR & A: RF & C (transfer) & B (own text) & All 8 & AUC \\
\midrule
HC-183 & 77.4$\pm$4.3 & 68--87 & 75.1 & 72.0 & 75.0$\pm$4.7 & 67.3$\pm$5.2 & 77.9$\pm$5.1 & 0.85 \\
Style-6 & 55.4$\pm$6.0 & 43--68 & 53.0 & 62.9 & 55.0$\pm$2.7 & 59.4$\pm$4.6 & 56.1$\pm$5.5 & 0.57 \\
STROKE & 60.8$\pm$5.2 & 49--72 & 59.8 & 60.1 & 53.9$\pm$4.0 & 54.2$\pm$5.8 & 61.1$\pm$5.1 & 0.62 \\
Combined & 75.5$\pm$3.9 & 64--86 & 73.8 & 66.4 & 70.2$\pm$4.5 & 61.7$\pm$5.3 & 70.0$\pm$4.5 & 0.82 \\
\addlinespace[2pt]
ResNet-18 & 69.5$\pm$4.6 & 58--81 & 68.7 & 60.8 & 63.7$\pm$3.9 & 53.1$\pm$4.5 & 65.6$\pm$4.0 & 0.74 \\
EffNet-B0 & 72.0$\pm$6.1 & 60--83 & 68.7 & 70.2 & 67.3$\pm$5.8 & 61.9$\pm$4.2 & 70.2$\pm$4.4 & 0.76 \\
ViT-B/16 & 70.4$\pm$3.4 & 58--82 & 68.7 & 71.5 & 66.5$\pm$3.6 & 65.6$\pm$5.0 & 71.7$\pm$4.3 & 0.74 \\
Stroke-CNN (frozen) & 65.8$\pm$6.3 & 55--77 & 65.7 & 66.8 & 64.2$\pm$3.3 & 53.7$\pm$4.2 & 64.6$\pm$4.9 & 0.70 \\
ResNet-18 (fine-tuned) & 71.5$\pm$5.2 & 60--83 & 71.5 & 70.5 & 72.0$\pm$5.5 & 70.6$\pm$4.3 & 70.6$\pm$4.4 & 0.79 \\
\addlinespace[2pt]
Photo int. & 66.4$\pm$2.3 & 52--79 & 67.1 & 66.7 & 61.4$\pm$2.6 & 66.1$\pm$2.4 & 69.4$\pm$3.0 & 0.75 \\
Paper only & 61.5$\pm$2.6 & 47--75 & 63.8 & 57.5 & 53.5$\pm$3.2 & 63.7$\pm$4.9 & 58.5$\pm$2.9 & 0.71 \\
\bottomrule
\end{tabular}
\end{table*}

\subsection{Photometric cues, time of photography and adjustment}
\label{sec:photo}
Photo intensity, seven grey-level statistics of the whole photograph, reached \vPhotoA\% (AUC \vPhotoAuc; $p=\vPhotoP$), although all pages were photographed with one phone in one session, and paper pixels alone reached \vPaperA\% ($p=\vPaperP$). Of the acquisition measures, only ink contrast differed: strokes of right-handed writers were darker relative to the paper (Hedges $g$\,=\,\vInkConG, $p=\vInkConP$). Paper brightness ($p=\vPaperLevP$), the share of the frame filled by text (\vBlockFracL{} vs.\ \vBlockFracR\%, $p=\vBlockFracP$), ink density ($p=\vInkDensP$) and page tilt (\vTiltL$^\circ$ vs.\ \vTiltR$^\circ$, $p=\vTiltP$) did not. The pseudo-pressure measures located the difference (Fig.~\ref{fig:dataset}c--f): right-handed writers' stroke centres were darker ($g=-0.68$, $p=0.029$), also at equal stroke width ($g=-0.75$, $p=0.017$; $q=0.086$ over six measures), whereas stroke edges ($p=0.37$) and stroke width ($p=0.52$) did not differ, and neither darkness measure followed the time of photography.

Lighting changed during the session of almost six hours: paper brightness and mean photo intensity rose with the time of photography in both groups (writer-level Spearman $\rho=0.59$ and $0.48$, $p\leq0.001$), but ink contrast did not ($\rho=0.11$, $p=0.49$), and capture time was only weakly related to class ($\rho=0.14$). With time regressed out, photo intensity kept 68.1\% ($p=0.011$) and HC-183 76.2\% ($p=0.001$; Table~\ref{tab:pit}): lighting drift does not explain the signal.

Regressing ink density and the seven photo-intensity statistics out of every feature inside the training folds lowered HC-183 from \vHcA\% to \vHcAdj\% ($p=0.027$, 999 permutations), Combined to \vCombAdj\% ($p=0.021$), the frozen ImageNet embeddings to \vDeepAdjRange\%, the frozen Stroke-CNN to \vScnnAdj\% and the fine-tuned ResNet-18 from \vGreyFtA\% to \vGreyFtAdj\% (interval \vGreyFtAdjLo--\vGreyFtAdjHi\%), and left STROKE at \vStrokeAdj\% ($p=\vStrokeAdjAllP$). After adjustment the gradient descriptor therefore remained the most accurate representation and the only significant one among the primary tests, while fine-tuning learned mainly what the adjustment removes. If stroke darkness reflects how the writers write (Section~\ref{sec:disc}), the adjusted values are lower bounds for handedness-related accuracy, not estimates of that accuracy. Removing the truncated images changed writer BA by \vTruncShiftMin{} to \vTruncShiftMax{} points, and re-including the folder excluded by the data preparer by \vOutlShiftMin{} to \vOutlShiftMax{} points.

\begin{table*}[t]
\caption{Inference, sensitivity and pitfall analyses (linear SVM, condition A, 20 repetitions; \%). $p$: whole-writer permutation. Adjusted, Time adj., Acq.+time: ink density and photo intensity, capture time (linear and quadratic), or both regressed out within training folds. Para.\ BA: per-image accuracy with writer-independent or image-level (leaky) folds. Writer ID: \vnWriters-way leave-one-paragraph-out 1-NN. C$-$D, B$-$E: text contrasts (points). n/a: grey levels regressed on themselves. Fine-tuned: writer ID averaged over 100 fold models; --: needs retraining per permutation or image fold.}
\label{tab:pit}
\centering\footnotesize
\setlength{\tabcolsep}{3.0pt}
\begin{tabular}{lccccccccc}
\toprule
 & Writer BA & $p$ & Adjusted & Time adj. & Acq.+time & Para.\ BA, writer & Para.\ BA, leaky & Writer ID & C$-$D\,/\,B$-$E \\
\midrule
HC-183 & 77.4 & 0.001 & 64.5 & 76.2 & 64.5 & 72.3 & 82.1 & 73.0 & +1.0 / $-$7.6 \\
Style-6 & 55.4 & 0.266 & 53.0 & 56.9 & 48.9 & 53.7 & 63.7 & 63.1 & +1.0 / +5.7 \\
STROKE & 60.8 & 0.090 & 60.0 & 60.5 & 55.4 & 59.1 & 94.2 & 100.0 & $-$6.2 / $-$5.8 \\
Combined & 75.5 & 0.002 & 64.0 & 74.0 & 61.5 & 72.8 & 97.0 & 98.0 & $-$4.6 / $-$12.3 \\
ResNet-18 & 69.5 & 0.003 & 52.0 & 70.5 & 53.3 & 67.2 & 91.1 & 94.4 & $-$5.1 / $-$10.9 \\
EffNet-B0 & 72.0 & 0.002 & 61.0 & 71.5 & 59.5 & 70.0 & 95.4 & 93.7 & $-$3.7 / $-$4.1 \\
ViT-B/16 & 70.4 & 0.004 & 56.2 & 68.6 & 55.4 & 68.1 & 95.3 & 94.4 & $-$2.3 / $-$2.0 \\
Stroke-CNN (frozen) & 65.8 & 0.020 & 51.1 & 71.1 & 52.9 & 63.5 & 84.9 & 86.1 & +0.4 / $-$10.2 \\
ResNet-18 (fine-tuned) & 71.5 & -- & 54.9 & 70.0 & 54.2 & 71.0 & -- & 96.3 & +0.7 / $-$0.6 \\
Photo int. & 66.4 & 0.025 & n/a & 68.1 & n/a & 68.6 & 71.6 & 21.4 & $-$5.0 / +2.1 \\
Paper only & 61.5 & 0.095 & n/a & 64.5 & n/a & 61.2 & 64.4 & 15.9 & $-$10.5 / +3.3 \\
\bottomrule
\end{tabular}
\end{table*}

\subsection{What the stroke features use}
\begin{figure*}[t]
\centering
\includegraphics[width=0.85\textwidth]{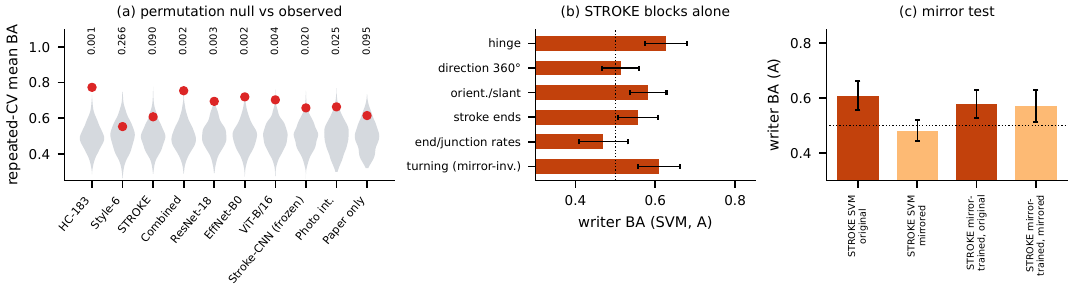}
\caption{(a) Permutation null distributions (999 whole-writer label permutations) of the repeated-CV mean BA; observed values in red, $p$ above. (b) Writer BA of each STROKE block alone (SVM, condition A). (c) Mirror test: the STROKE SVM applied to original and left--right mirrored test images, and the mirror-trained STROKE SVM fitted on original and mirrored training ink. Bars: mean$\pm$s.d.\ over repetitions.}
\label{fig:mech}
\end{figure*}

\begin{figure}[t]
\centering
\includegraphics[width=0.9\columnwidth]{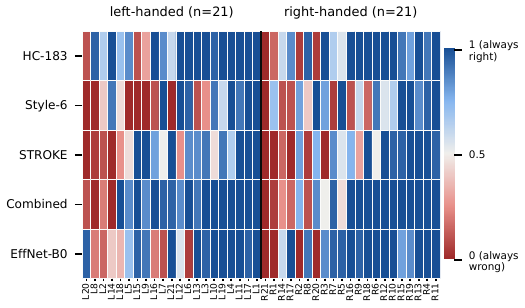}
\caption{Fraction of repetitions in which each writer (columns, sorted within class; L/R = left/right writer number) was classified correctly in condition A: red always wrong, grey chance, blue always right.}
\label{fig:writers}
\end{figure}

Fig.~\ref{fig:mech}b evaluates each STROKE block alone: contour hinge \vBlkHinge\%, turning \vBlkTurn\%, orientation and slant \vBlkOrient\% and stroke endings \vBlkEnd\%, whereas full-circle contour direction (\vBlkDir\%) and the endpoint/junction rates (\vBlkRates\%) were near chance. Applied to mirrored test images (Fig.~\ref{fig:mech}c), the STROKE SVM fell from \vStrokeMirOrig\% to \vStrokeMir\% (\vStrokeMirFlip\% of writers changed class), but this mirror effect (12.6 points) was not significant against 999 whole-writer permutations ($p=0.066$). A STROKE SVM trained on original and mirrored ink reached 57.9\% (mirrored test 57.3\%), so only about three points of STROKE's accuracy require the left--right direction of strokes; consistent with this, the mirror-invariant turning histogram alone reached \vBlkTurn\%. Per-writer errors overlapped little between STROKE and HC-183 (Spearman \vSpStrokeHc; Fig.~\ref{fig:writers}).

\subsection{Text conditions and evidence}
\begin{figure}[t]
\centering
\includegraphics[width=0.9\columnwidth]{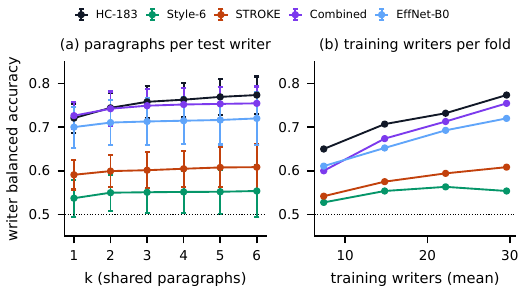}
\caption{(a) Writer BA when each test writer contributes $k$ randomly chosen shared paragraphs (200 draws per repetition). (b) Writer BA with 25, 50, 75 or 100\% of the training writers of each fold. Linear SVM, condition A; bars in (a) are s.d.\ over repetitions.}
\label{fig:curves}
\end{figure}

Table~\ref{tab:text} compares the conditions at matched evidence. Transfer to own texts did not cost HC-183 accuracy (C$-$D \vHcCminusD{} points) but changed STROKE by \vStrokeCminusD{}, and training on own texts was harder than on a shared pair (B$-$E \vHcBminusE{} and \vStrokeBminusE{}). More paragraphs per test writer raised HC-183 from \vEvHcOne\% to \vHcA\% (Fig.~\ref{fig:curves}a), and halving the training writers lowered it to \vLearnHHc\% (Fig.~\ref{fig:curves}b), so all representations remain limited by the number of writers.

\begin{table}[t]
\caption{Text conditions at matched evidence (writer BA, \%, linear SVM, 20 repetitions). D$_k$: model A tested on shared pair $k$ (1--2, 3--4, 5--6); C: model A tested on the writer-specific pair; E$_k$: trained and tested on shared pair $k$; B: trained and tested on the writer-specific pair.}
\label{tab:text}
\centering\scriptsize
\setlength{\tabcolsep}{2.1pt}
\begin{tabular}{lccccccccccc}
\toprule
 & A & D$_1$ & D$_2$ & D$_3$ & C & C$-$D & E$_1$ & E$_2$ & E$_3$ & B & B$-$E \\
\midrule
HC-183 & 77.4 & 71.7 & 73.0 & 77.3 & 75.0 & +1.0 & 68.1 & 69.9 & 86.7 & 67.3 & $-$7.6 \\
Style-6 & 55.4 & 55.8 & 49.2 & 57.0 & 55.0 & +1.0 & 52.9 & 52.1 & 56.2 & 59.4 & +5.7 \\
STROKE & 60.8 & 59.4 & 61.2 & 59.6 & 53.9 & $-$6.2 & 59.5 & 58.8 & 61.5 & 54.2 & $-$5.8 \\
Combined & 75.5 & 73.7 & 72.7 & 78.2 & 70.2 & $-$4.6 & 71.5 & 70.0 & 80.4 & 61.7 & $-$12.3 \\
\bottomrule
\end{tabular}
\end{table}

\subsection{What differs in the strokes}
\begin{figure*}[t]
\centering
\includegraphics[width=0.85\textwidth]{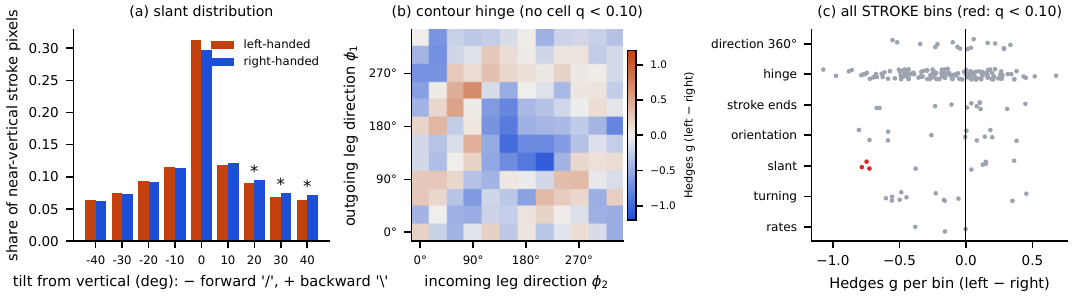}
\caption{Left--right differences of STROKE bins (writer means over the six shared paragraphs; post hoc, descriptive). (a) Slant of near-vertical strokes; *: $q<0.10$ within the block. (b) Hedges $g$ of the $12\times12$ contour-hinge cells (orange: more frequent in left-handed writers, blue: in right-handed writers). (c) Hedges $g$ of every bin by block; red: $q<0.10$.}
\label{fig:diff}
\end{figure*}

Fig.~\ref{fig:diff} compares the STROKE histograms of the two groups bin by bin (post hoc, descriptive). Nominal differences ($p<0.05$) appeared in slant, orientation, stroke endings and contour-hinge cells, but after false-discovery control within blocks only slant survived: \vSigSlant{} of \vNSlant{} slant bins had $q<0.10$, against \vSigHinge{} of \vNHinge{} hinge cells, \vSigOrient{} of \vNOrient{} orientation bins and no turning, direction or ending bin. Right-handed writers produced more near-vertical strokes leaning backwards, from upper left to lower right, by 15--45$^\circ$ (\vSlantBackR\% vs.\ \vSlantBackL\% of near-vertical stroke pixels; $g$ from \vSlantBackG, $q=\vSlantBackQ$). Page tilt did not differ, so the slant difference is not a rotation of the photographs.

\subsection{Evaluation pitfalls}
STROKE identified the writer of a held-out paragraph among \vnWriters{} with \vStrokeWid\% top-1 accuracy, the frozen ImageNet embeddings with \vDeepWidRange\% and HC-183 with \vHcWid\%. With a split over images instead of writers, the paragraph-level BA of STROKE rose from \vStrokeParaSplit\% to \vStrokeLeaky\% and that of the ImageNet embeddings from \vDeepParaRange\% to \vDeepLeakyRange\%: inflation that comes from recognising writers. Including the two unreleased raw folders as filed shifted writer BA by \vDupShiftMin{} to \vDupShiftMax{} points.

\section{Discussion}
\label{sec:disc}
Under controlled collection, handedness was identifiable for writers absent from training. With shared pens and paper, mixed writing sessions and a single photography session, the group difference in stroke darkness is unlikely to come from materials or photography. Lighting did drift during the session, but ink contrast did not follow it, and the signal survived adjustment for time. Darker strokes of right-handed writers are therefore best read as a writing trait: the extra ink sat in the centre of equally wide strokes, consistent with higher pen force \cite{franke2004ink}, although writing speed, pen angle and individual pens could produce a similar pattern. This cue explains part, not all, of the accuracy: with darkness and brightness removed, the gradient descriptor still reached \vHcAdj\%, whereas a ResNet-18 fine-tuned on the same photographs fell to \vGreyFtAdj\%. Because such a cue is invisible to permutation tests and easily mistaken for an acquisition artefact, studies should report photometric controls and adjusted estimates alongside raw accuracy.

The writing-only STROKE features exclude grey levels, ink amount and framing and were unaffected by the ink and brightness adjustment (\vStrokeAdj\%), but they reached only \vStrokeA\% (interval 49.2--71.8\%; $p=\vStrokeP$), and most of their accuracy did not require stroke direction. With \vnWriters{} writers, handedness is thus carried mainly by stroke darkness and gradient texture, and the evidence for a signal in stroke shape alone is weak. The bin-level difference that survived false-discovery control, more backward-leaning near-vertical strokes in right-handed writers, is plausible as an effect of hand position relative to the line, but needs confirmation on new writers.

The results suggest five rules for handedness and other soft-biometric studies: (i) split by writer and report writer-identification accuracy; (ii) report grey-level and paper-only controls and adjusted accuracy; (iii) photograph the groups in random, interleaved order with shared materials and record sessions; (iv) audit folders geometrically for re-photographed pages; and (v) use direction-sensitive diagnostics such as the mirror test. Comparisons with earlier handedness accuracies \cite{almaadeed2016handedness,rahmanian2021handedness} are difficult because protocols differ in whether writers are held out and acquisition is controlled.

The study has limits. \vnWriters{} writers give wide intervals, and the STROKE interval includes chance. The adjustment is linear, and its time version is post hoc. The groups were photographed in alternating blocks rather than in random order; the data come from one institution, one script and one phone; and earlier switching of the writing hand was not recorded. Pen pressure was inferred from ink deposition, not measured with a force sensor, and writing speed was not recorded. Next steps are measuring pressure with a digitising tablet and testing the frozen pipelines on new writers.

\section{Conclusion}
In a writer-independent evaluation of 42 writers whose Bangla paragraphs were written with shared materials and photographed under identical conditions, a gradient descriptor and frozen ImageNet embeddings identified the handedness of unseen writers with up to \vHcA\% balanced accuracy (interval 67.7--86.5\%). Part of this signal is stroke darkness, consistent with higher writing pressure in right-handed writers and not explained by photography; without it the descriptor still reached \vHcAdj\%. Writing-only stroke shape was a weak cue (\vStrokeA\%), with right-handed writers producing more backward-leaning strokes. Reporting photometric controls and adjusted estimates allows such cues to be interpreted rather than dismissed as artefacts.

\section*{Data and code availability}
The BanglaHand dataset is available at Mendeley Data (\href{https://doi.org/10.17632/n4sjstdzfy.1}{doi:10.17632/n4sjstdzfy.1}) under a CC BY 4.0 licence \cite{banglahand2026data}. The analysis code, saved results and permutation null distributions are available at \url{https://github.com/LegendaryBeast/banglahand-handedness-} under the MIT licence.

\section*{CRediT authorship contribution statement}
\textbf{Tanzim Hasan Prappo:} Conceptualization, Methodology, Software, Validation, Writing -- original draft, Writing -- review \& editing. \textbf{Sayeed Hassan, Sayem Ali, Sukhendra Nath Dhar:} Investigation, Validation. \textbf{M. Shahidur Rahman:} Supervision, Writing -- review \& editing.

\section*{Ethics, funding and competing interests}
Informed consent was obtained voluntarily from all participants prior to data collection. No personal identifying information was collected, and all handwriting samples are pseudonymous. This research received no specific grant; writers received an honorarium paid by the authors from personal funds. The authors declare no competing interests.

\section*{Declaration of generative AI use}
During the preparation of this work the authors used an AI assistant to audit the data, validate code and results. The authors reviewed and edited all content and take full responsibility for the publication.

\end{document}